\documentclass[10pt,twocolumn,letterpaper]{article}

\usepackage{algorithmicx}
\usepackage{algorithm}
\usepackage{algpseudocode}
\usepackage{iccv}
\usepackage{times}
\usepackage{epsfig}
\usepackage{graphicx}
\usepackage{amsmath}
\usepackage{amssymb}
\usepackage{dsfont}
\usepackage{bm}

\usepackage[pagebackref=true,breaklinks=true,letterpaper=true,colorlinks,bookmarks=false]{hyperref}

\iccvfinalcopy 

\def\iccvPaperID{4180} 

\ificcvfinal\fi

\begin{document}

\setlength{\textfloatsep}{6pt}
\setlength{\intextsep}{6pt}

\title{Instance Segmentation in 3D Scenes using Semantic Superpoint Tree Networks}

\author{Zhihao Liang$^{1,2}$,
Zhihao Li\textsuperscript{3}, Songcen Xu\textsuperscript{3},
Mingkui Tan\textsuperscript{1} and Kui Jia$^{1,4,5}$\thanks{Correspondence to Kui Jia $<$kuijia@scut.edu.cn$>$}\\
\textsuperscript{1}South China University of Technology,\quad \textsuperscript{2}DexForce Technology Co., Ltd. \\
\textsuperscript{3}Noah's Ark Lab, Huawei Technologies,\quad \textsuperscript{4}Pazhou Laboratory,\quad \textsuperscript{5} Peng Cheng Laboratory\\
{\tt\small  eezhihaoliang@mail.scut.edu.cn},
{\tt\small \{kuijia, mingkuitan\}@scut.edu.cn}, \\
{\tt\small \{zhihao.li, xusongcen\}@huawei.com}
}

\maketitle
\ificcvfinal\thispagestyle{empty}\fi

\begin{abstract}
   Instance segmentation in 3D scenes is fundamental in many applications of scene understanding. It is yet challenging due to the compound factors of data irregularity and uncertainty in the numbers of instances. State-of-the-art methods largely rely on a general pipeline that first learns point-wise features discriminative at semantic and instance levels, followed by a separate step of point grouping for proposing object instances. While promising, they have the shortcomings that (1) the second step is not supervised by the main objective of instance segmentation, and (2) their point-wise feature learning and grouping are less effective to deal with data irregularities, possibly resulting in fragmented segmentations. To address these issues, we propose in this work an end-to-end solution of Semantic Superpoint Tree Network (SSTNet) for proposing object instances from scene points. Key in SSTNet is an intermediate, semantic superpoint tree (SST), which is constructed based on the learned semantic features of superpoints, and which will be traversed and split at intermediate tree nodes for proposals of object instances. We also design in SSTNet a refinement module, termed CliqueNet, to prune superpoints that may be wrongly grouped into instance proposals. Experiments on the benchmarks of ScanNet and S3DIS show the efficacy of our proposed method. At the time of submission, SSTNet ranks top on the ScanNet (V2) leaderboard, with 2\% higher of mAP than the second best method. The source code in PyTorch is available at \url{https://github.com/Gorilla-Lab-SCUT/SSTNet}.
\end{abstract}

\vspace{-0.5cm}

\section{Introduction}
\label{SecIntro}

\begin{figure}[htbp]
    \centering
    \includegraphics[width=0.4\textwidth]{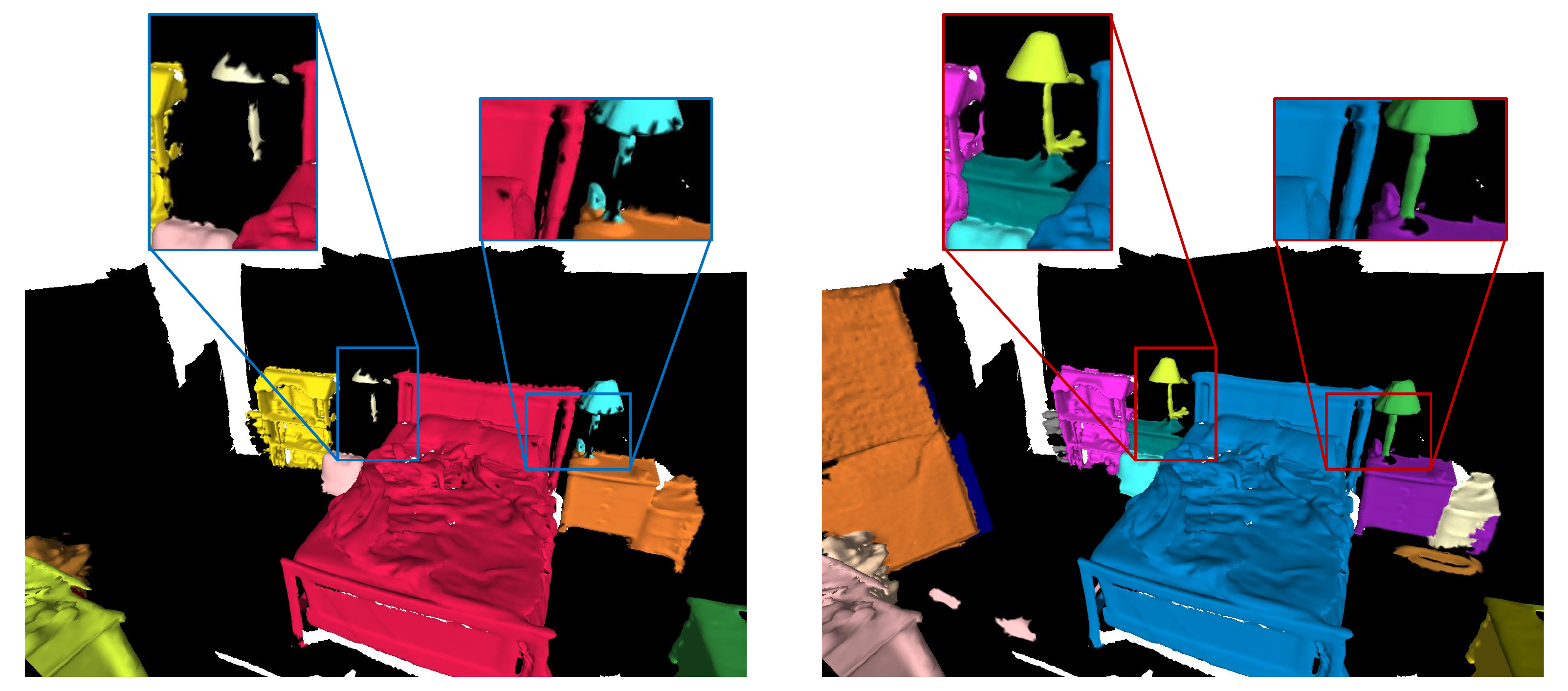}
    \caption{Visualization of example instance segmentation results from an existing, point-wise grouping method (PointGroup\cite{jiang2020pointgroup}, left) and our SSTNet (right). Different colors represent segmented instances.}
    \label{figFragmented}
\end{figure}

The task of 3D instance segmentation is fundamental in many applications concerned with 3D scene understanding. Given an observed scene of point cloud reconstructed from depth cameras
via multi-view fusion techniques \cite{dai2017bundlefusion, Han2018FlashFusionRG}, the task is to both assign semantic labels of pre-defined object categories to individual scene points, and differentiate those belonging to different object instances. Learning to achieve 3D instance segmentation is challenging at least in the following aspects: (1) observed scene points are usually sparse and irregular, which poses difficulties for learning point-wise classification based on shape features of local (and possibly global) contexts around individual points; (2) the unknown number of object instances in a scene introduces additional uncertainties to the problem of learning point-instance associations that is already combinatorial; (3) even though point-wise classification and point-instance associations can be conducted, learning consistencies among spatially adjacent points are not guaranteed, which may cause fragmented segmentations, especially around object boundaries (cf. Fig. \ref{figFragmented} for an illustration).

State-of-the-art methods \cite{jiang2020pointgroup,zhang2019point,engelmann20203d}, e.g., those ranking top on the ScanNet benchmark \cite{dai2017scannet}, tackle (some of) the above challenges with the following general pipeline. They first train networks to learn point-wise features that are discriminative at both the semantic and instance levels, followed by a separate step of point clustering that groups together those believed to be on same instances, using the learned point-wise features. While promising, they have the following shortcomings. Firstly, the second step of point clustering is independent of network training, whose results are thus not guaranteed by guiding towards the ground-truth groupings of object instances. Secondly, while superpoints \cite{landrieu2018large} have been commonly used for semantic segmentation of 3D points \cite{landrieu2019point, cheng2020cascaded}, when coming to instance segmentation, these state-of-the-art methods, except OccuSeg \cite{han2020occuseg}, choose to conduct both feature learning and grouping in a point-wise manner, which takes away their chance to leverage the geometric regularities established at the mid-level shape representation of superpoints.  

\begin{figure*}
    \centering
    \includegraphics[width=0.85\textwidth]{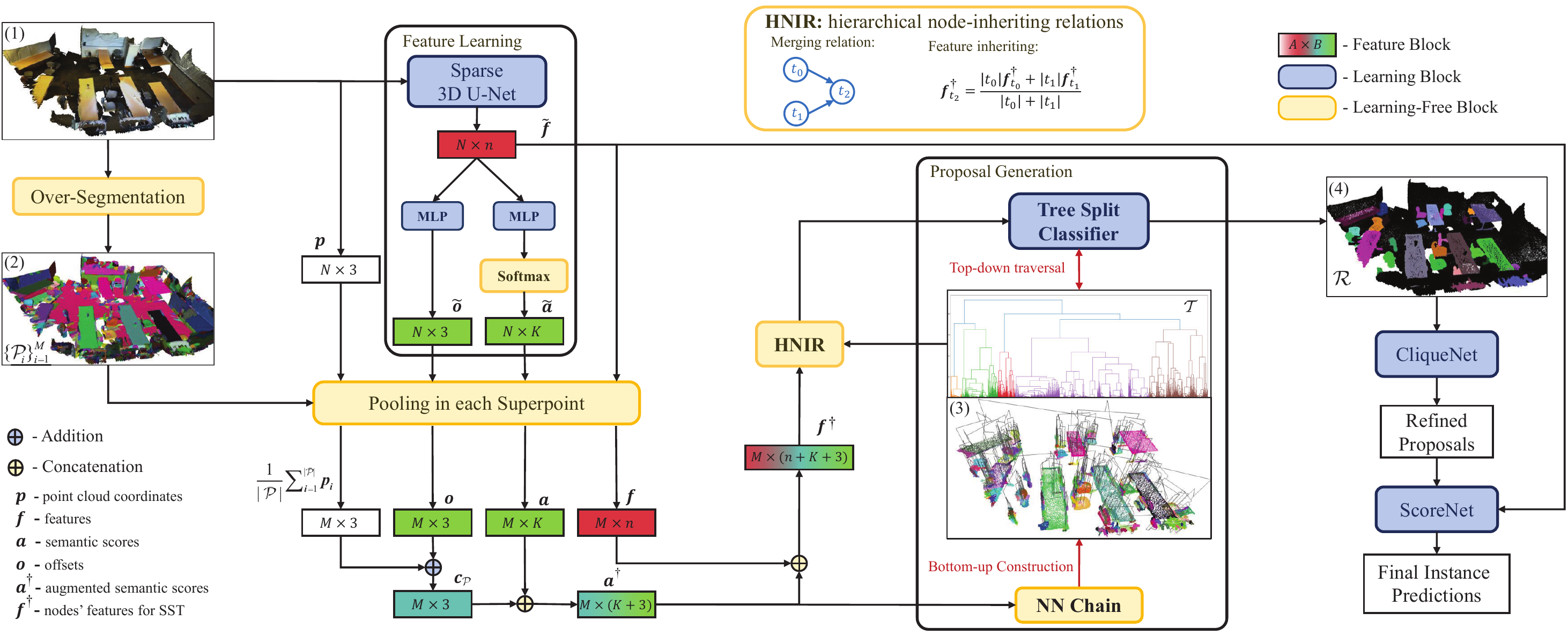}
    \caption{Overview of our proposed Semantic Superpoint Tree Network (SSTNet). Please refer to the main text for details of the individual modules. $N$ is the number of scene points, $M$ is the number of superpoints, $K$ is the number of categories, and $n$ is the dimension of output features from the backbone. $\tilde{\bm{f}}, \tilde{\bm{a}}, \tilde{\bm{o}}$ denote the point-wise features, semantic scores and offsets respectively. (1) input scene, (2) generated superpoint set $\{\mathcal{P}_i\}^M_{i=1}$, (3) foreground superpoints and Semantic Superpoint Tree(SST)$\mathcal{T}$, (4) generated proposals $\mathcal{R}$ after tree traversal and splitting. Nearest-neighbor chain (NN Chain) is the algorithm we use for efficiently constructing the tree. }
\label{figArchitecture}
\end{figure*}

To overcome these shortcomings, we are motivated to develop an end-to-end solution for proposing object instances from an observed scene of points. Considering that a superpoint represents a geometrically homogeneous neighborhood, we choose to work with superpoints pre-computed from the scene points, and the problem of instance segmentation boils down as learning a network that groups superpoints on same object instances. In this work, we design such a solution called \emph{Semantic Superpoint Tree Network (SSTNet)}, as illustrated in Fig. \ref{figArchitecture}. Similar to existing methods, SSTNet starts with a backbone that learns point-wise semantic and instance-level features; differently from them, SSTNet immediately aggregates these features as superpoint-wise ones efficiently via point-wise pooling. Key in SSTNet is an intermediate, semantic superpoint tree (SST), with the superpoints as its tree leaves. SST is constructed based on the pooled semantic (and instance-level) features of superpoints, and will be traversed and split by the subsequent SSTNet module of binary classification; starting from the root, a proposal of object instance is formed as the superpoints of a tree branch when non-splitting decision is made at the intermediate tree node that spans the branch (cf. Fig. \ref{figTree} for an illustration). Our tree construction is highly efficient by choosing ways of feature inheritance from leaves to the root and pair-wise similarity metric, which support fast algorithms such as nearest-neighbor chain \cite{willett1987multidimensional}. We note that erroneous assignments of superpoints to instances may occur when constructing and traversing the tree. To compensate, we design a subsequent refinement module termed CliqueNet, which converts each proposed branch as a graph clique and learns to prune some of the branch nodes. A ScoreNet \cite{jiang2020pointgroup} is finally used to evaluate the generated proposals, which gives instance segmentation results of our SSTNet. 

Thorough experiments on the benchmark datasets of ScanNet \cite{dai2017scannet} and S3DIS\cite{armeni20163d} show the efficacy of our proposed method. Notably, SSTNet outperforms all existing methods on the two benchmarks, and at the time of submission, it ranks top on the ScanNet (V2) leaderboard, with 2\% higher of mAP than the second best method. We finally summarize our technical contributions as follows.

\begin{itemize}
\item We propose an end-to-end solution of \emph{Semantic Superpoint Tree Network (SSTNet)} to directly propose and evaluate object instances from observed 3D scenes. By working with superpoints, our method enjoys the benefit of geometric regularity that supports consistent and sharp segmentations, especially at object boundaries.

\item We choose a strategy of divisive grouping in SSTNet, which first builds the tree, followed by tree traversal for object proposal via node splitting. By constructing the tree with appropriate node merging and feature inheritance, our strategy is an order of magnitude faster than the alternative, agglomerative grouping, thus enabling efficient training and inference of SSTNet.

\item Considering that erroneous assignments of superpoints to instances may occur when constructing and traversing the tree, we design a refinement module in SSTNet, termed CliqueNet, which converts each proposed branch as a graph clique and learns to prune some of the branch nodes. Experiments show its efficacy. 
\end{itemize}

\section{Related Works}
\label{secRelatedWork}

In this section, we briefly review the literature of 3D segmentation, focusing on those relevant to elements of our proposed method.

\noindent\textbf{3D Semantic Segmentation}
Establishing geometric regularities is essential to realize semantic segmentation for an irregular point cloud.
Recent methods used projection\cite{lawin2017deep}, voxelization\cite{Graham_2018_CVPR, choy20194d} or local aggregation\cite{NIPS2017_d8bf84be, wang2019dynamic} to perform brief regularization, while the subsequent semantic learning task is still challenging.
Instead, superpoint-based\cite{landrieu2018large, landrieu2019point} methods aggregated the geometrically homogeneous points as superpoints to establish a certain degree of geometric regularities.
Furthermore, superpoints become the mid-level shape representation to boild down the problem of instance segmentation as grouping the superpoints that belong to the same instance.

\noindent\textbf{3D Instance Segmentation}
Considering bottom-up methods, which cluster results based on semantic segmentation. \cite{zhang2020ssen, wang2019associatively, lahoud20193d} heuristically\cite{comaniciu2002mean, inproceedings} clustered instance masks based on discriminative instance-level features\cite{debrabandere2017semantic}. Intuitively, PointGroup\cite{jiang2020pointgroup} utilized the adjacency of instance-wise coordinates.
The above clustering results relied on the boundary conditions due to the lack of explicit boundary supervision.
To address this issue, SSTNet combines the bottom-up clustering strategy with top-down traversal to realize end-to-end learning proposal generation.

\noindent\textbf{Image Segmentation for Object Proposals}
To overcome the complexity caused by sliding windows\cite{Girshick_2014_CVPR, ren2015faster}, segmentation-based\cite{2019Hierarchical, Rantalankila_2014_CVPR, Wang_2015_CVPR} methods treat 2D detection as image segmentation, where the candidates are hypothesized from hierarchical image segmentation using an agglomeration manner.
Furthermore, SSTNet involves a greedy agglomeration strategy and employs a learning splitting classifier to get rid of dependence on the times of agglomeration and generate precise mask results.

\section{Overview}
\label{secOverview}

Assume an input point set $\mathcal{I} = \{\bm{p} \in \mathbb{R}^3 \}$ of reconstructed 3D scene from depth cameras via multi-view fusion techniques (e.g., SLAM \cite{dai2017bundlefusion, Han2018FlashFusionRG}), 
which contains an unknown number of object instances of $K$ categories. The task is to segment out those points in $\mathcal{I}$ that define each of such instances, and is challenging as analyzed in Section \ref{SecIntro}.
To alleviate the difficulty, we choose to establish a certain degree of geometric regularities for points in $\mathcal{I}$, by leveraging a mid-level shape representation called superpoints \cite{landrieu2018large,landrieu2019point} --- for an input $\mathcal{I}$, superpoints define geometrically homogeneous neighborhoods of its local points, and are usually computed by over-segmenting $\mathcal{I}$ using graph partition. \footnote{By using superpoints, we rely on the assumption that individual superpoints would not be across object boundaries; while this is not guaranteed, it is the cost that we would like to trade for the benefit of geometric regularities which the superpoints bring to the original, irregular point set $\mathcal{I}$.} With the set of superpoints $\{ \mathcal{P} \}$  pre-computed from $\mathcal{I}$, the problem of instance segmentation boils down as grouping together spatially close $\{ \mathcal{P} \}$ that belong to a same object instance and assigning them a semantic label.
This is technically a clustering/grouping problem in the 3D space where superpoints live; given spatial compactness for superpoints on a same instance, it is natural to consider hierarchical clustering/grouping to achieve the goal. The strategy resembles those used for object proposals in 2D images via hierarchical image segmentation \cite{2019Hierarchical, Ren_2013_CVPR, guimaraes2012hierarchical}. To implement the above idea, we propose in this work an end-to-end, hierarchical segmentation network that is trained to semantically group superpoints of a scene as object instances of pre-defined categories. The key in our network is an intermediate, semantic superpoint tree (SST); it is constructed based on the learned semantic features in the preceding network module, and will be traversed and split in the subsequent network module; we thus term our proposed method as Semantic Superpoint Tree Network (SSTNet). Fig. \ref{figArchitecture} gives the illustration.

\vspace{-0.2cm}
More specifically, SSTNet starts with a \textbf{backbone} that learns point-wise feature $\tilde{\bm{f}} \in \mathbb{R}^{n}$ for each $\bm{p} \in \mathcal{I}$, which is then fed into a subsequent module of \textbf{semantic scoring} to output semantic score $\tilde{\bm{a}} \in [0, 1]^K$ and offset $\tilde{\bm{o}} \in \mathbb{R}^3$, where $\tilde{\bm{a}}$ is a $K$-dimensional probability vector representing soft label prediction of the point $\bm{p}$, and $\tilde{\bm{o}}$ is a predicted coordinate offset relative to center of the instance to which $\bm{p}$ belongs. In parallel with this module we apply over-segmentation to $\mathcal{I}$ to have the superpoints $\{ \mathcal{P} \}$; note that this is applied only once during network training. The point-wise $\{ \tilde{\bm{f}} \}$, $\{ \tilde{\bm{a}} \}$, and $\{ \tilde{\bm{o}} \}$ are aggregated, via average pooling, inside each superpoint to form the superpoint feature $\bm{f} \in \mathbb{R}^{n}$, score $\bm{a} \in [0, 1]^K$, and offset $\bm{o} \in \mathbb{R}^3$. Assume that a collection of superpoints are obtained from $\mathcal{I}$, we use the thus obtained $\{ \bm{f} \}$, $\{ \bm{a} \}$, and $\{ \bm{o} \}$ for use in subsequent modules of the network. To achieve efficient training of SSTNet, we choose divisive grouping (i.e., in a top-down manner) after \textbf{construction of semantic superpoint tree $\mathcal{T}$}, instead of agglomerative grouping commonly used in hierarchical image segmentation \cite{Ren_2013_CVPR, 2019Hierarchical}, which means that the whole tree $\mathcal{T}$ is first constructed whose leaf nodes represent individual superpoints. We then design a module of \textbf{tree traversal and splitting} that learns to hierarchically split the tree nodes; starting from the root, a proposal of object instance is formed as a tree branch when non-splitting decision is made at an intermediate tree node. We note that erroneous assignments between superpoints and object instances may occur during both stages of tree construction and tree traversal and splitting. \cite{pan2019deep, Nie_2020_CVPR} demonstrated the refinement can achieve higher accuracy for mesh reconstruction. Inspired by them, we design a subsequent refinement module termed \textbf{CliqueNet} to compensate for some of these errors. This module converts each proposal branch as a graph clique and learns to prune some of the branch nodes.
We finally use a ScoreNet \cite{jiang2020pointgroup} to evaluate the generated proposals, which gives instance segmentation results of our SSTNet. The whole network is trained in an end-to-end manner, which, to the best of our knowledge, is the first one for the task of 3D instance segmentation on point set. The intermediate SST construction is highly efficient, whose computational complexity and running time are given in Section \ref{SecSSTConstruction}. Section \ref{SecModules} also presents individual modules of the network and compares with alternative designs.

\section{Individual Modules of the Proposed Network}
\label{SecModules}

\subsection{Backbone and Semantic Scoring}

Assume that the input $\mathcal{I}$ contains $N$ points. Given $\{\bm{p}_i \in \mathcal{I} \}_{i=1}^N$, we use a 3D convolutional backbone of U-Net style \cite{ronneberger2015u} to learn the point-wise features $\{\tilde{\bm{f}}_i \in \mathbb{R}^n \}_{i=1}^N$, whose layers are implemented as submanifold sparse convolution (SSC) or sparse convolution (SC) \cite{Graham_2018_CVPR}. We give layer specifics in the supplementary material.

We obtain the semantic scoring $\{\tilde{\bm{a}}_i \in [0, 1]^K \}_{i=1}^N$ and offset prediction $\{\tilde{\bm{o}}_i \in \mathbb{R}^3 \}_{i=1}^N$ from $\{\tilde{\bm{f}}_i \}_{i=1}^N$, by employing two multi-layer perceptrons (MLPs) respectively. Let $\{\tilde{\bm{a}}_i^{*} \}_{i=1}^N$ denote the ground-truth semantic labels of the $N$ points in the form of $K$-dimensional, one-hot vector. We use the following loss to train the MLP for semantic scoring
\vspace{-0.5cm}

\begin{eqnarray}\label{EqnSemanticScoring}
L_{\textrm{\tiny semantic}} = - \frac{1}{N}\sum_{i=1}^N \texttt{CE}(\tilde{\bm{a}}_i, \tilde{\bm{a}}_i^{*}) + \qquad\qquad\qquad \nonumber \\  1 - \frac{2\sum_{i=1}^N \tilde{\bm{a}}_i^{\top} \tilde{\bm{a}}_i^{*}}{\sum_{i=1}^N \tilde{\bm{a}}_i^{\top} \tilde{\bm{a}}_i  + \sum_{i=1}^N \tilde{\bm{a}}_i^{*\top} \tilde{\bm{a}}_i^{*}} ,
\end{eqnarray}
where $\texttt{CE}(\cdot, \cdot)$ denotes the cross-entropy loss, and the remaining terms in (\ref{EqnSemanticScoring}) define a dice loss that alleviates the imbalance among the $K$ categories \cite{milletari2016v}. Let $\bm{c}_{\bm{p}}^{*}$ denote the geometric center of the object instance to which any $\bm{p} \in \mathcal{I}$ belongs. We use the following loss to train the MLP for offset prediction
\vspace{-0.2cm}
\begin{eqnarray}\label{EqnOffset}
L_{\textrm{\tiny offset}} =  \frac{1}{N'} \sum_{i=1}^N \| \tilde{\bm{o}}_i - (\bm{c}_{\bm{p}_i}^{*} - \bm{p}_i) \|_2 \cdot \mathds{I}(\bm{p}_i) - \qquad \nonumber \\ \frac{1}{N'} \sum_{i=1}^N \frac{\tilde{\bm{o}}_i^{\top}}{\|\tilde{\bm{o}}_i\|_2} \cdot \frac{\bm{c}_{\bm{p}_i}^{*} - \bm{p}_i}{\| \bm{c}_{\bm{p}_i}^{*} - \bm{p}_i \|_2} \cdot \mathds{I}(\bm{p}_i) ,
\end{eqnarray}
where $\mathds{I}(\bm{p}) \in \{0, 1\}$ is an indicator function telling whether the point $\bm{p}$ belongs to any object instance, and $N' = \sum_{i=1}^N \mathds{I}(\bm{p}_i)$ counts the number of such points. We give specifics of the two MLPs in the supplementary material.

\subsection{Construction of Semantic Superpoint Tree}
\label{SecSSTConstruction}

As stated in the preceding section, our construction of SST $\mathcal{T}$ is based on superpoints $\{ \mathcal{P} \}$ pre-computed from the input $\mathcal{I}$; without loss of generality, we assume $M$ ones are computed from $\mathcal{I}$. Features $\{ \bm{f}_i \in \mathbb{R}^n \}_{i=1}^M$, semantic scores $\{ \bm{a}_i \in [0, 1]^K \}_{i=1}^M$, and offsets $\{ \bm{o}_i \in \mathbb{R}^3 \}_{i=1}^M$ at the superpoint level are obtained simply via average pooling over those point-wise ones inside each of superpoints $\{ \mathcal{P}_i \}_{i=1}^M$.

\begin{figure}[htbp]
    \centering
    \includegraphics[width=0.4\textwidth]{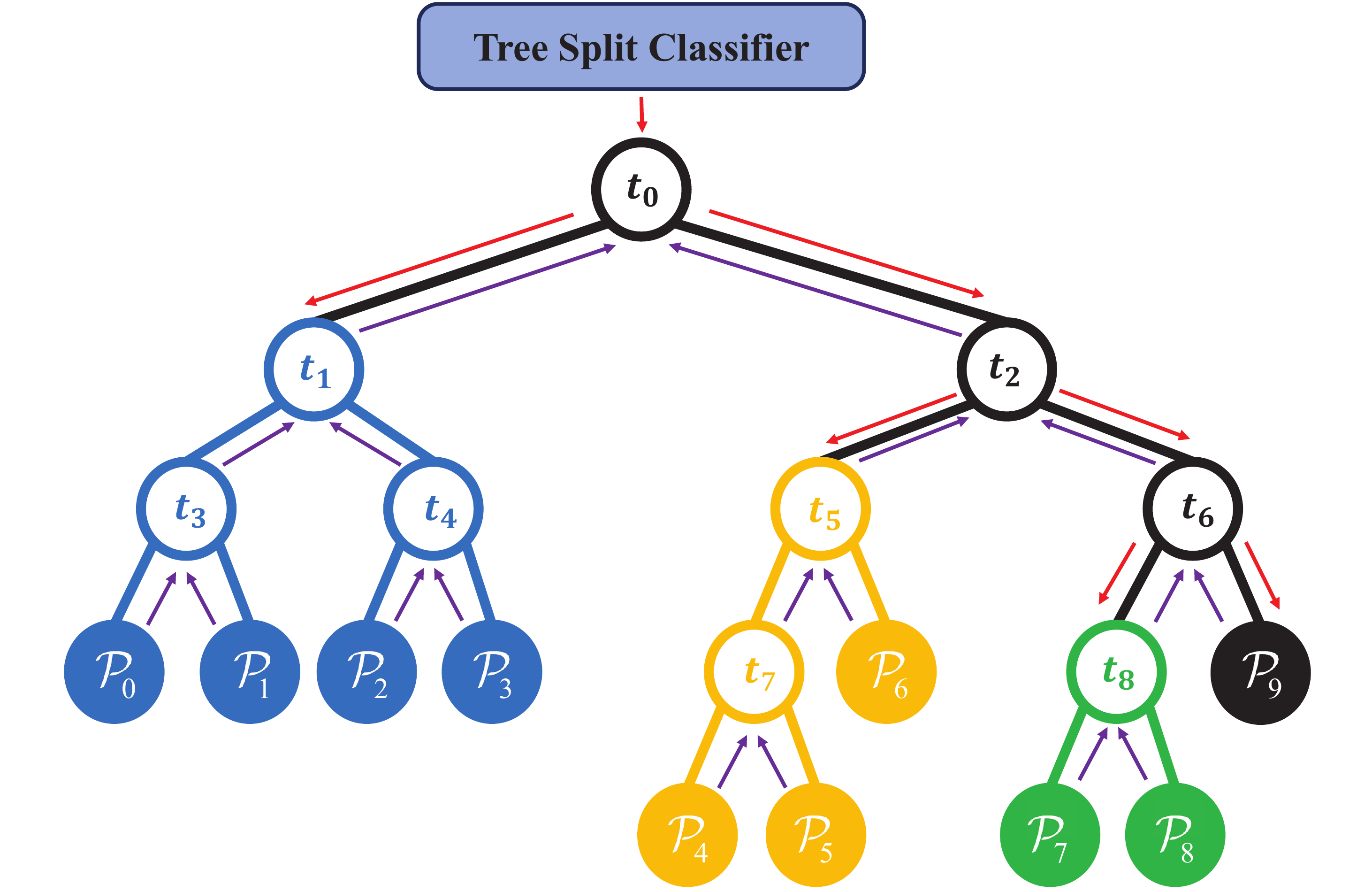}
    \caption{Illustration on the construction and traversal of semantic superpoint tree (SST). {\color{purple}{$\rightarrow$}} represents the bottom-up construction process; {\color{red}{$\rightarrow$}} represents the top-down traversal process.}
    \label{figTree}
\end{figure}

Given the predicted $\{\bm{f}_i, \bm{a}_i, \bm{o}_i\}_{i=1}^M$ for $\{ \mathcal{P}_i \}_{i=1}^M$, a tree can grow greedily \cite{mullner2011modern}, starting from merging the leaf nodes of superpoints (cf. Fig. \ref{figTree} for an illustration). To define the linkage criteria, there exist many choices of similarity metric between any pair of $\mathcal{P}_i$ and $\mathcal{P}_j$. In this work, we choose semantic score and offset prediction over the triple $\{\bm{f}, \bm{a}, \bm{o}\}$ to define the metric. Specifically, for a superpoint $\mathcal{P}$, we first compute the predicted geometric center of a (possible) object instance to which it may belong as $\bm{c}_{\mathcal{P}} = \bm{o} + \frac{1}{|\mathcal{P}|}\sum_{i=1}^{|\mathcal{P}|} \bm{p}_i$, and then concatenate $\bm{a}^{\dag} = [\bm{a}; \bm{c}_{\mathcal{P}}] \in \mathbb{R}^{K+3}$ \footnote{Considering the domain difference of $\bm{a} \in [0, 1]^K$ and $\bm{c}_{\mathcal{P}} \in \mathbb{R}^3$, we ever try weighted concatenation such as $\bm{a}^{\dag} = [\alpha \bm{a}; \beta \bm{c}_{\mathcal{P}}]$, where $\alpha$ and $\beta$ are hyper-parameters. We end with the empirical setting of $\alpha = \beta = 1$, which gives good results in practice. }. We use the augmented $\bm{a}_i^{\dag}$ and $\bm{a}_j^{\dag}$ to represent $\mathcal{P}_i$ and $\mathcal{P}_j$, and compute the Euclidean distance $ \| \bm{a}_i^{\dag} - \bm{a}_j^{\dag} \|$ as the linkage criterion that determines the ordering of pair-wise superpoint merging. Merging two superpoints $\mathcal{P}_i$ and $\mathcal{P}_j$ results in an intermediate tree node, denoted as $t \in \mathcal{T}$. We compute semantic score of $t$, via weighted averaging, as
\vspace{-0.2cm}
\begin{eqnarray}\label{EqnFeaInheritage}
\bm{a}_t = w_i \bm{a}_i + w_j \bm{a}_j ,
\end{eqnarray}
where the weights $w_i$ and $w_j$ are proportional to the respective sizes of $\mathcal{P}_i$ and $\mathcal{P}_j$, i.e., $w_i = |\mathcal{P}_i| / (|\mathcal{P}_i| + |\mathcal{P}_j|)$ and $w_j = |\mathcal{P}_j| / (|\mathcal{P}_i| + |\mathcal{P}_j|)$. Offset prediction of $t$ is computed similarly as $\bm{o}_t = w_i \bm{o}_i + w_j \bm{o}_j$. We then compute the augmented $\bm{a}_t^{\dag}$ from the obtained $\bm{a}_t$ and $\bm{o}_t$. Note that we also compute the feature $\bm{f}_t = w_i \bm{f}_i + w_j \bm{f}_j$ for the node $t$, which will be used in the subsequent module of proposal generation via tree traversal and splitting. Given the augmented $\bm{a}_t^{\dag}$ for any $t \in \mathcal{T}$ and the pair-wise similarity metric based on Euclidean distance, the tree can be constructed hierarchically, as illustrated in Fig. \ref{figTree}, whose depth ranges between $\log_2^M$ and $M-1$. For clarity, we write the $M$ leaf nodes as $\{ t_{\mathcal{P}_i} \in \mathcal{T} \}_{i=1}^M$ and any root or intermediate one as $t \in \mathcal{T}$.

Our use of the augmented semantic score $\bm{a}^{\dag} = [\bm{a}; \bm{c}_{\mathcal{P}}]$ to represent each $t_{\mathcal{P}}$ (and $t$) is based on the argument that for any pair of $\mathcal{P}_i$ and $\mathcal{P}_j$ on a same instance, both their semantic scores and instance centers are expected to be consistent. Empirical results in Section \ref{subsecAblationStudies} show that it gives better performance than alternative choices do, which verifies the hypothesis. We thus term the constructed $\mathcal{T}$ as semantic superpoint tree.

\noindent\textbf{Remarks} Given the $M$ superpoints, the hierarchical tree construction described above has a complexity of $\mathcal{O}(M^3)$. Due to the linear feature inheritance (\ref{EqnFeaInheritage}) and the use of Euclidean distance as the similarity metric, construction of $\mathcal{T}$ can be made highly efficient by employing the fast algorithm of nearest-neighbor chain \cite{mullner2011modern}, which results in a same $\mathcal{T}$ at a complexity of $\mathcal{O}(M^2)$. On a machine running at 13 Hz, it takes $\sim$ 75 milliseconds per construction (e.g., scenes of ScanNet \cite{dai2017scannet}), thus supporting online SST construction per iteration of network training.

\subsection{Proposal Generation via Tree Traversal and Splitting}
\label{SecProposalGenViaTreeTraversal}

Given the constructed SST $\mathcal{T}$, our proposed SSTNet generates proposals of object instance by learning a binary classifier that traverses and splits nodes of $\mathcal{T}$. For any root or intermediate node $t$, denote its two child nodes as $s_1 \in \mathcal{T}$ and $s_2 \in \mathcal{T}$. Each $t$ in fact defines a tree branch, denoted as $\mathcal{B}_t$, that contains leaf nodes of superpoints. As stated in Section \ref{SecSSTConstruction}, feature $\bm{f}_t$ and augmented score $\bm{a}_t^{\dag}$ associated with each $t$ have been hierarchically inherited from its contained superpoints. We use the concatenated $\bm{f}_t^{\dag} = [\bm{f}_t; \bm{a}_t^{\dag}] \in \mathbb{R}^{n + K + 3}$ as feature of node $t$.

Denote the binary classifier to be learned as $\phi: \mathbb{R}^{n + K + 3} \times \mathbb{R}^{n + K + 3} \in (0, 1)$. Starting from the root node, we maintain a queue of tree traversal in a breadth-first manner. Let $\mathcal{Q}$ and $\mathcal{R}$ be two empty sets, and push the root into the queue $\mathcal{Q}$. A node $t$ is to be split once $\phi(\bm{f}_{s_1}^{\dag}, \bm{f}_{s_2}^{\dag}) < 0.5$ , i.e., the two child nodes of $t$ are believed to belong to different object instances; we then push $s_1$ and $s_2$ into the queue $\mathcal{Q}$. Conversely, when $\phi(\bm{f}_{s_1}^{\dag}, \bm{f}_{s_2}^{\dag}) \geq 0.5$, we consider all superpoints contained in the tree branch $\mathcal{B}_t$ as a proposal of object instance, and push $t$ into $\mathcal{R}$; we stop traversing the intermediate nodes contained in $\mathcal{B}_t$. Note that we have established an index table of the hierarchical node-inheriting relations when constructing $\mathcal{T}$, which supports efficient retrieval of both intermediate and leaf nodes/superpoints contained in any branch $\mathcal{B}_t$. All proposals of object instance would be obtained in $\mathcal{R}$ when the queue $\mathcal{Q}$ becomes empty. Algorithm \ref{algTraversalProposalsGeneration} gives pseudo code of the above procedure.

In this work, we implement the classifier $\phi$ as an MLP, whose details are given in the supplementary material. To train $\phi$, we define the instance-level, ground-truth labels for nodes of the tree as follows. Assume that a training scene $\mathcal{I}$ contains $J$ object instances, which may belong to some of the $K$ categories. For any superpoint $\mathcal{P}$ (i.e, a leaf node $t_{\mathcal{P}}$), we assign its instance-level, soft label $\bm{q}_{\mathcal{P}}^{*} \in [0, 1]^J$ according to what proportions its contained points belong to (some of) the $J$ instances. The soft label $\bm{q}_t^{*} \in [0, 1]^J$ for any intermediate or root $t$ is again hierarchically inherited, via weighted averaging, from those of superpoints, similar to the inheritance of features. Given that $s_1$ and $s_2$ are the two child nodes of $t$ in $\mathcal{T}$, we use the following loss symmetric to them to train $\phi$
\vspace{-0.2cm}
\begin{eqnarray}\label{EqmSplittingLoss}
L_{\textrm{\tiny splitting}} = \mathbb{E}_{t \in \mathcal{T}/\{ t_{\mathcal{P}_i} \}_{i=1}^M }  \frac{1}{2} [ \texttt{BCE}(\phi(\bm{f}_{s_1}^{\dag}, \bm{f}_{s_2}^{\dag}), \bm{q}_{s_1}^{*\top}\bm{q}_{s_2}^{*}) + \nonumber \\ \texttt{BCE}(\phi(\bm{f}_{s_2}^{\dag}, \bm{f}_{s_1}^{\dag}), \bm{q}_{s_1}^{*\top}\bm{q}_{s_2}^{*}) ] ,
\end{eqnarray}

where $\texttt{BCE}(\cdot, \cdot)$ denotes a binary cross-entropy loss, and $\bm{q}_{s_1}^{*\top}\bm{q}_{s_2}^{*} \in [0, 1]$ indicates, in a soft manner, whether the two child nodes belong to a same instance.

\noindent\textbf{Remarks} In the proposed SSTNet, we choose to first build the tree, as described in Section \ref{SecSSTConstruction}, and then learn to traverse and split tree nodes to generate instance proposals; in other words, we choose a strategy of divisive grouping, instead of an agglomerative one commonly used in hierarchical image segmentation \cite{2019Hierarchical, Ren_2013_CVPR, guimaraes2012hierarchical}. Our motivation for such a design is mostly computational: by using nearest-neighbor chain \cite{mullner2011modern}, our tree construction has a complexity of $\mathcal{O}(M^2)$, and the tree traversal to propose all the branches of object instances has a complexity of $\mathcal{O}(M)$, giving rise to an overall complexity of $\mathcal{O}(M^2 + M)$; in contrast, learning to generate proposals in an agglomerative manner has an order-of-magnitude higher complexity of $\mathcal{O}(M^3)$.

\renewcommand{\algorithmicrequire}{ \textbf{Input: }} 
\renewcommand{\algorithmicensure}{ \textbf{Output:}} 
\begin{algorithm}[htb]
   \caption{Pseudo code of proposal generation via tree traversal and splitting}
   \label{algTraversalProposalsGeneration}
   \begin{algorithmic}[1]
     \Require tree $\mathcal{T}$, node features $\{\bm{f}^\dagger_i\}_{i=1}^{|\mathcal{T}|}$, classifier $\phi$;
     \State initialize $\mathcal{R} = \emptyset$ to store proposals, and queue $\mathcal{Q} = \emptyset$;
     \State push the root of $\mathcal{T}$ into $\mathcal{Q}$;
     \While{- $\mathcal{Q}$.\texttt{isempty}() }
         \State $t=\mathcal{Q}$.dequeue()
         \If{- $t$.\texttt{isleaf}()}
             \State $\{s_1, s_2\}$ = $t$.\texttt{getchild}() 
             \State $\bm{f}^\dagger_{s_1}$ = $s_1$.\texttt{getfeature}()
             \State $\bm{f}^\dagger_{s_2}$ = $s_2$.\texttt{getfeature}()
             \If{$\phi(\bm{f}_{s_1}^{\dag}, \bm{f}_{s_2}^{\dag}) \geq 0.5$}
                 \State push $t$ into $\mathcal{R}$, and $\mathcal{B}_t$ = $t$.\texttt{getbranch}()
             \Else
                 \State $\mathcal{Q}$.enqueue($s_1, s_2$)
             \EndIf
         \EndIf
     \EndWhile
     \State \Return $\mathcal{R}$;
   \end{algorithmic}
 \end{algorithm}

\subsection{CliqueNet for Refinement of Proposals}

We note that in the forward pass of SSTNet, once a superpoint $\mathcal{P}$ truly on an object instance is constructed into a wrong branch $\mathcal{B}_t$ of SST $\mathcal{T}$, e.g., $\mathcal{B}_t$ corresponding to the background or a different instance, the mistake cannot be corrected. Nevertheless, when any branch $\mathcal{B}_t$ is proposed as an object instance, we have the chance to improve its score evaluation (cf. Section \ref{SecScoreNet}) by pruning its contained superpoints that may belong to other instances or the background.

\begin{figure}[htbp]
    \centering
    \includegraphics[width=0.35\textwidth]{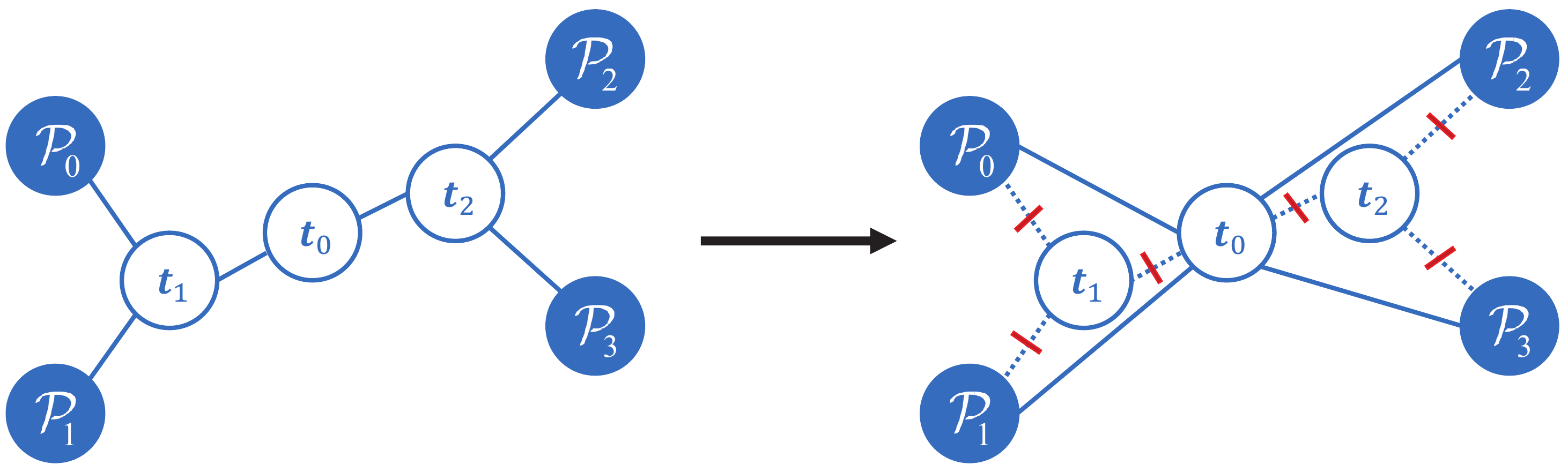}
    \caption{Illustration on conversion of a tree branch as a (graph) clique.}
    \label{figRefinement}
\end{figure}

Consider a proposed branch $\mathcal{B}_t$ consisting of $M_t$ leaf nodes of superpoints.  A straightforward way to implement the pruning is to concatenate feature representation $\bm{f}_t^{\dag}$ at node $t$ with each $\bm{f}_{\mathcal{P}}^{\dag}$ of $\{ \bm{f}_{\mathcal{P}_i}^{\dag} \}_{i=1}^{M_t}$, and to learn a binary classifier that decides whether the superpoint $\mathcal{P}$ should be removed. This, however, involves only the pair-wise relation between $\bm{f}_t^{\dag}$ and each $\bm{f}_{\mathcal{P}}^{\dag}$, and we empirically find that it is less effective to prune erroneously assigned superpoints. In this work, we propose a more effective scheme, termed CliqueNet, that determines which superpoints to remove by learning the feature interactions among $\{\bm{f}_t^{\dag}, \bm{f}_{\mathcal{P}_1}^{\dag}, \dots, \bm{f}_{\mathcal{P}_{M_t}}^{\dag} \}$. Specifically, given the proposed branch $\mathcal{B}_t$ as shown in Fig. \ref{figRefinement}, we first connect the node $t$ directly with individual leaf nodes/superpoints, which forms a clique $\mathcal{C}$ when thinking of the whole SST $\mathcal{T}$ as a graph --- we note that the cliques formed for different proposed branches are independent with each other, i.e., they are not on a same graph. An adjacency matrix $\bm{A}_{\mathcal{C}} \in \{0, 1\}^{(M_t+1) \times (M_t+1)} $ can be computed that specifies node connections of the clique. Let $\bar{\bm{A}}_{\mathcal{C}} = \bm{A}_{\mathcal{C}} + \bm{I}$, where $\bm{I}$ is an identity matrix, and write features of clique nodes compactly as $\bm{F}_{\mathcal{C}}^{\dag} = [\bm{f}_t^{\dag}, \bm{f}_{\mathcal{P}_1}^{\dag}, \dots, \bm{f}_{\mathcal{P}_{M_t}}^{\dag}] \in \mathbb{R}^{(n+K+3)\times (M_t + 1)}$. Denote the CliqueNet as a function $\psi$, the first layer of $\psi$ computes
\begin{eqnarray}\label{EqmCliqueNetLayerComp}
\texttt{ReLU}( \bar{\bm{D}}_{\mathcal{C}}^{-1/2} \bar{\bm{A}}_{\mathcal{C}} \bar{\bm{D}}_{\mathcal{C}}^{-1/2} \bm{F}_{\mathcal{C}}^{\dag} \bm{W}_{\psi}^1) ,
\end{eqnarray}
where $\bar{\bm{D}}_{\mathcal{C}}$ is the diagonal degree matrix of $\bar{\bm{A}}_{\mathcal{C}}$, and $\bm{W}_{\psi}^1$ denotes weight matrix of the first layer of $\psi$. In this work, we use a three-layer CliqueNet whose specifics are given in the supplementary material.

CliqueNet outputs scores $\psi(\bm{F}_{\mathcal{C}}^{\dag}, \bm{A}_{\mathcal{C}}) \in (0, 1)^{M_t + 1}$ defined respectively for the $M_t + 1$ nodes in $\mathcal{C}$. To train $\psi$, we impose supervision on each node pair of $t$ and $\mathcal{P}_i$, $i \in \{1, \dots, M_t\}$, giving rise to
\begin{eqnarray}\label{EqmCliqueNetLoss}
L_{\textrm{\tiny refining}} =  \frac{1}{M_t} \sum_{i=1}^{M_t} \texttt{BCE}(\psi(\bm{F}_{\mathcal{C}}^{\dag}, \bm{A}_{\mathcal{C}}), \bm{q}_t^{*\top}\bm{q}_{\mathcal{P}_i}^{*} ) ,
\end{eqnarray}
where the instance-level, soft labels $\bm{q}_t^{*} \in [0, 1]^J$ and $\bm{q}_{\mathcal{P}}^{*} \in [0, 1]^J$ are defined in Section \ref{SecProposalGenViaTreeTraversal}.

\begin{figure*}
    \centering
    \includegraphics[width=0.95\textwidth]{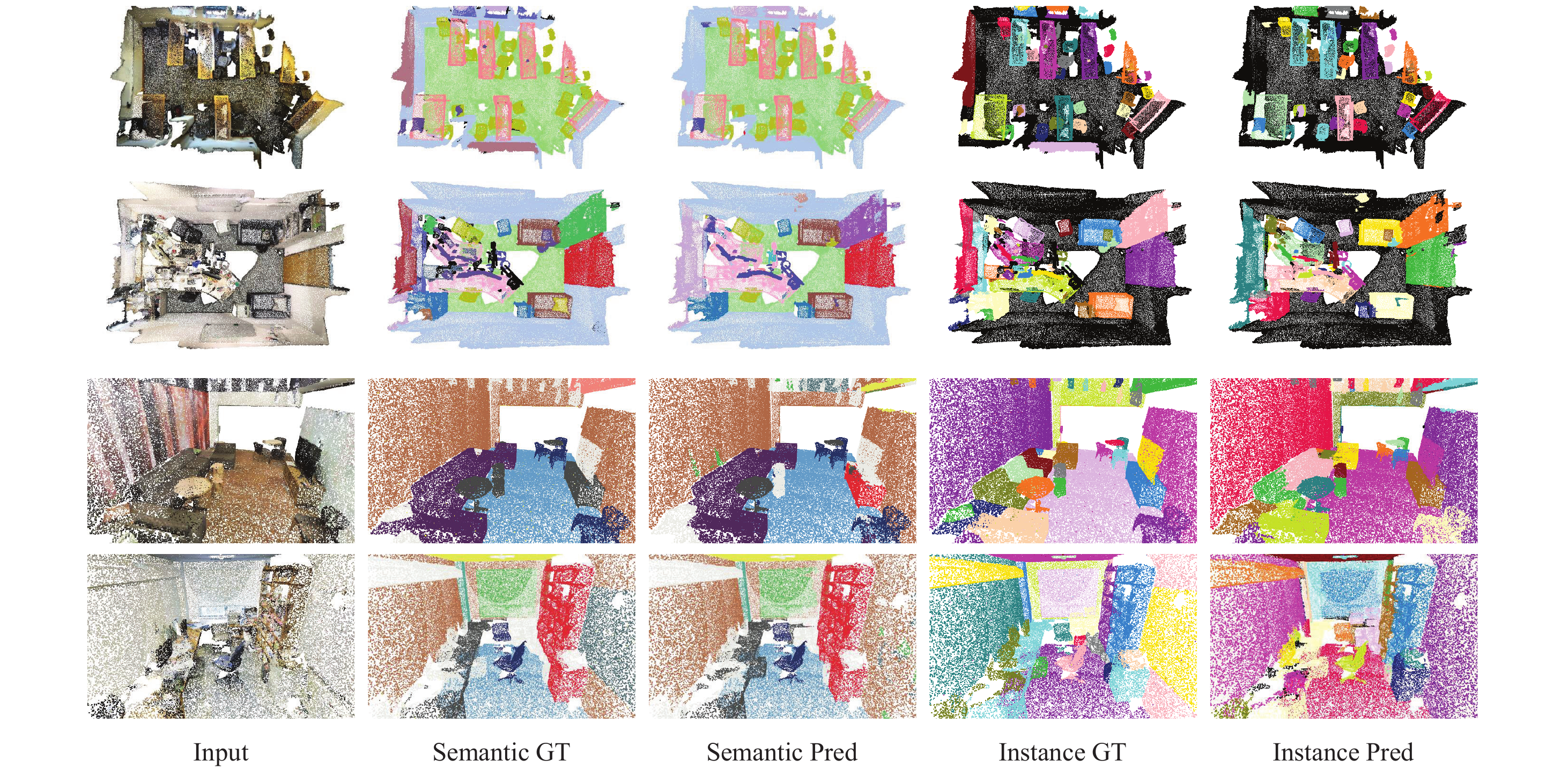}
    \caption{Visualization of the semantic and instance segmentation results on the validation set of ScanNet v2 (top) and S3DIS (bottom).}
    \label{figVisualization}
\end{figure*}

\subsection{Proposal Evaluation}
\label{SecScoreNet}

Denote a proposed branch of object instance, after pruning some superpoints by CliqueNet, as $\mathcal{B}_t^{-}$, and assume that it contains $N_t^{-}$ raw points. Recall that their point-wise features have been computed by the backbone of SSTNet. We write these features compactly as $\widetilde{\bm{F}}_{\mathcal{B}_t^{-}} = [\tilde{\bm{f}}_1, \dots, \tilde{\bm{f}}_{N_t^{-}}] \in \mathbb{R}^{n\times N_t^{-}}$. We follow \cite{jiang2020pointgroup} and use a ScoreNet, denoted as $\omega$, to evaluate the proposal. The ScoreNet is simply a miniature of U-Net; one may refer to \cite{jiang2020pointgroup} for the network details. Depending on the intersection-over-union (IoU) value with the ground-truth instances in the scene $\mathcal{I}$, we define label of the proposal as $v_t^{*} \in [0, 1]$ (cf. the supplementary material for details of setting the $v_t^{*}$ value), and train the ScoreNet with the following loss
\begin{eqnarray}\label{EqmEvaluationLoss}
L_{\textrm{\tiny evaluation}} = \frac{1}{|\mathcal{R}|}\sum_{t \in \mathcal{R}}\texttt{BCE}(\omega(\widetilde{\bm{F}}_{\mathcal{B}_t^{-}}), v_t^{*}),
\end{eqnarray}
where $|\mathcal{R}|$ is the number of proposals generated by our SSTNet (cf. Algorithm \ref{algTraversalProposalsGeneration}).

\subsection{Training and Inference}
\label{SecTrainInference}

We write our overall objective for training SSTNet as
\begin{eqnarray}\label{OverallLoss}
L_{\textrm{\tiny SSTNet}} = L_{\textrm{\tiny semantic}} + L_{\textrm{\tiny offset}} + L_{\textrm{\tiny splitting}} + L_{\textrm{\tiny refining}} + L_{\textrm{\tiny evaluation}} .
\end{eqnarray}
Note that SSTNet is trained in a greedy, module-wise manner, which means that the individual loss terms applied to their respective modules are sequentially invoked into the overall loss (\ref{OverallLoss}). Although the tree $\mathcal{T}$ needs to be constructed in every forward pass of SSTNet, it is highly efficient as indicated by the complexity and practical running time given in preceding sections. The complexity of tree traversal for instance proposals is linear w.r.t. the number of superpoints; furthermore, once a proposal is formed at an intermediate tree node, it is not necessary to traverse all the descendant nodes. The inference is simply a same procedure as a forward pass of SSTNet training. Given the non-overlapping nature of our proposed object instances, post-processing steps such as non-maximum suppression are not necessary.

\section{Experiments}
\label{secExperiments}

\begin{table*}[!htb]
    \resizebox{\textwidth}{!}{
        \begin{tabular}{c|c|cccccccccccccccccc}
        \hline
        Method & \textbf{AP} & bath & bed & bkshf & cab & chair & cntr & curt & desk & door & ofurn & pic & fridg & showr & sink & sofa & table & toilet & wind\\
        \hline
        3D-MPA\cite{engelmann20203d} & 35.5 & 45.7 & 48.4 & 29.9 & 27.7 & 59.1 & 4.7 & 33.2 & 21.2 & 21.7 & 27.8 & 19.3 & 41.3 & 41.0 & 19.5 & 57.4 & 35.2 & 84.9 & 21.3 \\
        SSEN\cite{zhang2020ssen} & 38.4 & \textbf{85.2} & 49.4 & 19.2 & 22.6 & 64.8 & 2.2 & 39.8 & 29.9 & 27.7 & 31.7 & 23.1 & 19.4 & 51.4 & 19.6 & 58.6 & 44.4 & 84.3 & 18.4 \\
        PE\cite{zhang2019point} & 39.6 & 66.7 & 46.7 & 44.6 & 24.3 & 62.4 & 2.2 & \textbf{57.7} & 10.6 & 21.9 & 34.0 & 23.9 & 48.7 & 47.5 & 22.5 & 54.1 & 35.0 & 81.8 & 27.3 \\
        PointGroup\cite{jiang2020pointgroup} & 40.7 & 63.9 & 49.6 & 41.5 & 24.3 & 64.5 & 2.1 & 57.0 & 11.4 & 21.1 & 35.9 & 21.7 & 42.8 & 66.0 & 25.6 & 56.2 & 34.1 & 86.0 & \textbf{29.1} \\
        OccuSeg\cite{han2020occuseg} & 48.6 & 80.2 & 53.6 & 42.8 & \textbf{36.9} & \textbf{70.2} & \textbf{20.5} & 33.1 & \textbf{30.1} & \textbf{37.9} & \textbf{47.4} & 32.7 & 43.7 & \textbf{86.2} & 48.5 & 60.1 & 39.4 & 84.6 & 27.3\\
        \hline
        \textbf{Our SSTNet} & \textbf{50.6} & 73.8 & \textbf{54.9} & \textbf{49.7} & 31.6 & 69.3 & 17.8 & 37.7 & 19.8 & 33.0 & 46.3 & \textbf{57.6} & \textbf{51.5} & 85.7 & \textbf{49.4} & \textbf{63.7} & \textbf{45.7} & \textbf{94.3} & 29.0 \\
        \hline
        \end{tabular}
    }
    \caption{3D instance segmentation on ScanNet (V2) benchmark (hidden testing set). Results of SSTNet are obtained by submitting onto the testing server the model trained on the ScanNet training set on January 4th, 2021. \label{tableBenchmark}}
\end{table*}

\noindent{\textbf{Datasets}} We conduct experiments using the benchmark datasets of ScanNet (V2) \cite{dai2017scannet} and S3DIS\cite{armeni20163d}. ScanNet has 1201 training, 312 validation, and 100 test scenes that contain object instances of 18 categories. Surface normals are also provided for each scene. We do analysis and ablation studies on its validation set, and submit our results to the hidden test set. S3DIS contains 6 large-scale indoor scenes with 13 object classes, we evaluate our model in the following aspects: (1) Area-5 is treated as the testing, while residuals are used for training, and (2) 6-fold cross validation that each area is treated as the testing once.
 
\begin{table}
    \centering
    \scalebox{0.9}{
    \begin{tabular}{c|ccc}
    \hline
    Method & mAP & AP@50 & AP@25\\
    \hline
    3D-MPA\cite{engelmann20203d} & 35.5 & 61.1 & 73.7 \\
    SSEN\cite{zhang2020ssen} & 38.4 & 57.5 & 72.4 \\
    PE\cite{zhang2019point}  & 39.6 & 64.5 & 77.6 \\
    PointGroup\cite{jiang2020pointgroup} & 40.7 & 63.6 & 77.8 \\
    OccuSeg\cite{han2020occuseg} & 48.6 & 67.2 & 74.2 \\
    \textbf{Our SSTNet} & \textbf{50.6} & \textbf{69.8} & \textbf{78.9} \\
    \hline
    \end{tabular}
    }
    \caption{3D instance segmentation on ScanNet (V2) benchmark (hidden testing set). Results of SSTNet are obtained by submitting onto the testing server the model trained on the ScanNet training set on January 4th, 2021. \label{tableScanNetResultsAP25AP50}}
\end{table}

\noindent\textbf{Implementation Details} For each input scene, we concatenate the RGB values and point coordinates as the point-wise inputs of SSTNet. The network is trained using AdamW optimizer \cite{loshchilov2019decoupled}, with an initial learning rate of 1e-3 and weight decay of 1e-4; learning rates follow a polynomial learning rate policy. We set the batch size as 4. We pre-process scenes of S3DIS dataset by sub-sampling its points at a rate of 1/4. We employ a graph-based segmentation method \cite{felzenszwalb2004efficient} to generate superpoints for ScanNet scenes. For S3DIS, each scene is represented by colored point cloud and we employ SPP + SPG \cite{landrieu2018large,landrieu2019point} to generate its superpoints. Module and layer specifics of SSTNet are given in the supplementary material. 

\noindent\textbf{Evaluation Metrics}
Following the official ScanNet (V2) evaluation protocol, we report mean Average Precisions (mAPs) at different thresholds of IoU as the evaluation metric to compare different methods. The mAP@25 and mAP@50 denote the average precision scores with IoU thresholds respectively set to 25\% and 50\%, and the mAP averages the scores with IoU thresholds set from 50\% to 95\%, with a step size of 5\%.

\subsection{Ablation Studies and Analyses}
\label{subsecAblationStudies}
We first conduct ablation studies to evaluate the efficacy of individual components proposed in SSTNet. These studies are conducted on the ScanNet (V2) dataset \cite{dai2017scannet}.

\noindent\textbf{Analysis on Features for SST Construction} The quality of SST depends on what features are used when contructing the tree. In this work, for a superpoint $\mathcal{P}$, we choose semantic score $\bm{a}$ and predicted instance center $\bm{c}_{\mathcal{P}}$ over the triple $\{\bm{f}, \bm{a}, \bm{c}_{\mathcal{P}}\}$, where $\bm{c}_{\mathcal{P}}$ is computed from the offset prediction $\bm{o}$ (cf. Section \ref{SecSSTConstruction}), and form the augmented $\bm{a}^{\dag} = [\bm{a}; \bm{c}_{\mathcal{P}}]$ for SST construction. Results in Table \ref{tableAblationSimilarityMetric} verify our argument that for any pair of superpoints on a same instance, their semantic scores and instance centers are expected to be consistent, while their superpoint-wise features are not necessarily to be similar.

\begin{table}[!htb]
    \centering
    \scalebox{0.8}{
    \begin{tabular}{ccc|ccc}
    \hline
    {\footnotesize Superpoint} & {\footnotesize Semantic} & {\footnotesize Instance} & {\footnotesize mAP} & {\footnotesize AP@50} & {\footnotesize AP@25} \\
    {\footnotesize feature} & {\footnotesize score} & {\footnotesize center} &   &  &  \\
    \hline
    \checkmark & & & 40.1 & 55.3 & 66.2 \\
    & \checkmark & & 43.5 & 59.8 & 72.2 \\
    & & \checkmark & 47.3 & 61.6 & 71.4 \\
    \checkmark & \checkmark & \checkmark & 48.9 & 63.6 & 72.9 \\
    & \checkmark & \checkmark & \textbf{49.4} & \textbf{64.3} & \textbf{74.0} \\
    \hline
    \end{tabular}
    }
    \caption{Analysis on the features used for SST construction. Experiments are conducted on the validation set of ScanNet (V2) \cite{dai2017scannet}. Refer to Section \ref{SecSSTConstruction} for how the three types of features are computed. \label{tableAblationSimilarityMetric}}
\end{table}

\noindent\textbf{Efficacy of Proposal Generation via Tree Traversal and Split Learning} To verify the efficacy of our main proposal generation scheme via SST, we compare with two alternatives. The first alternative conducts the same traversal of SST but replaces the node-splitting classifier $\phi$ with a simple thresholding scheme, which we term as \emph{SST-Thresholding}; to determine whether an intermediate node $t$ is to be split into its child nodes $s_1$ and $s_2$, it thresholds the Euclidean distance $\| \bm{a}_{s_1}^{\dag} - \bm{a}_{s_2}^{\dag} \|_2$ where we optimally tune the thresholds for its best performance \footnote{We also try thresholding of the Euclidean distance $\| \bm{f}_{s_1}^{\dag} - \bm{f}_{s_2}^{\dag} \|_2$, where $\bm{f}_{s_1}^{\dag} = [\bm{f}_{s_1}; \bm{a}_{s_1}^{\dag}]$ and $\bm{f}_{s_2}^{\dag}$ is computed similarly. It empirically gives even worse performance. }. For the second alternative, instead of relying on SST construction, given the $M$ superpoints with its augmented semantic scores $\{ \bm{a}_i^{\dag} \}_{i=1}^M$, we first build a K-nearest-neighbor graph based on pair-wise Euclidean distances, and then train a classifier to decide whether some graph edges should be disconnected; the resulting, disconnected graph cliques are proposed as object instances; we term this alternative as \emph{Superpoint Graph}, which can be interpreted as a flattened version of learning to propose object proposals. Table \ref{tableAblationLearningSplitting} shows that SST-thresholding performs the best at the low-precision metric of mAP@25, suggesting our construction of SST is indeed useful for generation of object proposals. On the averaged metric of mAP, SSTNet greatly outperforms the two alternatives.

\noindent\textbf{Efficacy of the CliqueNet Refinement} Ablation study on the efficacy of CliqueNet is presented in Table \ref{tableAblationCliqueNet}, which shows that pruning superpoints from proposed tree branches is effective at high-precision regimes of mAP metrics.

\subsection{Results on the ScanNet Benchmark}
\label{subsecScanNetBenchmark}
We train SSTNet on the training set of  ScanNet (V2) and submit our model onto the testing sever. Table \ref{tableBenchmark} shows that on the leaderboard of ScanNet (V2) test set, SSTNet outperforms all existing methods. Results at the metrics of AP@25 and AP@50 are reported in Table \ref{tableScanNetResultsAP25AP50}.
\vspace{-0.1cm}

\subsection{Results on S3DIS}
\label{subsecS3DIS}

Following the protocols used in previous methods, we employ the Area-5 and 6-fold cross validation, and use the mAP/AP@50/mean precision (mPrec)/mean recall(mRec) with IoU threshold 0.5 to evaluate SSTNet on the S3DIS dataset. One may refer to \cite{wang2019associatively} for precise definitions of these metrics. Table \ref{tableS3DIS} shows that SSTNet outperforms all exist methods, confirming the generalizable advantage of our proposed method.

\begin{table}
    \centering
    \scalebox{0.8}{
    \begin{tabular}{c|ccc}
    \hline
    Method & mAP & AP@50 & AP@25\\
    \hline
    SST-Thresholding & 46.3 & 62.6 & \textbf{74.7} \\
    Superpoint Graph & 44.4 & 60.6 & 69.5 \\
    Our SSTNet & \textbf{49.4} & \textbf{64.3} & 74.0 \\
    \hline
    \end{tabular}
    }
    \caption{Analyses on the efficacy of our proposal generation via traversal and node-splitting learning of semantic superpoint tree. Experiments are conducted on the validation set of ScanNet (V2) \cite{dai2017scannet}. Reer to the main text for how the two alternatives are designed.  \label{tableAblationLearningSplitting}}
\end{table}

\begin{table}
    \centering
    \scalebox{0.8}{
    \begin{tabular}{c|ccc}
    \hline
    CliqueNet Refining & mAP & AP@50 & AP@25\\
    \hline
    & 49.4 & 64.3 & \textbf{74.0} \\
    \checkmark & \textbf{50.0} & \textbf{64.7} & 73.9 \\
    \hline
    \end{tabular}
    }
    \caption{Ablation study on the efficacy of CliqueNet for proposal refinement. Experiments are conducted on the validation set of ScanNet (V2) \cite{dai2017scannet}. \label{tableAblationCliqueNet}}
\end{table}

\begin{table}
    \centering
    \scalebox{0.9}{
    \begin{tabular}{c|cccc}
    \hline
    Method & mAP & AP@50 & mPrec & mRec \\
    \hline
    ASIS\cite{wang2019associatively} & - & - & 55.3 & 42.4 \\
    PointGroup\cite{jiang2020pointgroup} & - & 57.8 & 61.9 & 62.1 \\
    $\textbf{Our SSTNet}$ & \textbf{42.7} & \textbf{59.3} & \textbf{65.5} & \textbf{64.2} \\
    \hline
    $\text{ASIS}^\dag$\cite{wang2019associatively} & - & - & 63.6 & 47.5 \\
    $\text{3D-BoNet}^\dag$\cite{yang2019learning} & - & - & 65.6 & 47.6 \\
    $\text{OccuSeg}^\dag$\cite{han2020occuseg} & - & - & 72.8 & 60.3 \\
    $\text{PointGroup}^\dag$\cite{jiang2020pointgroup} & - & 64.0 & 69.6 & 69.2 \\
    $\textbf{Our SSTNet}^\dag$ & \textbf{54.1} & \textbf{67.8} & \textbf{73.5} & \textbf{73.4} \\
    \hline
    \end{tabular}
    }
    \caption{Results of instance segmentation on the S3DIS validation set. Methods without the $\dag$ marks are evaluated on Area-5; methods marked with $\dag$ are evaluated on 6-fold cross validation. \label{tableS3DIS}}
\end{table}

\noindent\textbf{Acknowledgement} This work was partially supported by the Guangdong R$\&$D key project of China (No.: 2019B010155001), the National Natural Science Foundation of China (No.: 61771201), and the Program for Guangdong Introducing Innovative and Entrepreneurial Teams (No.: 2017ZT07X183).

{\small
\bibliographystyle{ieee_fullname}
\bibliography{egbib}
}

\end{document}